\documentclass[10pt,twocolumn,letterpaper]{article}

\usepackage{cvpr}
\usepackage{times}
\usepackage{epsfig}
\usepackage{graphicx}
\usepackage{amsmath}
\usepackage{amssymb}
\usepackage{dsfont}
\usepackage{subfigure}
\usepackage{balance}
\usepackage{multirow}

\usepackage[pagebackref=true,breaklinks=true,letterpaper=true,colorlinks,bookmarks=false]{hyperref}

\usepackage{colortbl}

\usepackage{booktabs,xcolor,siunitx}
\definecolor{lightgray}{gray}{0.9}

\definecolor{lBlue}{rgb}{0.65,0.8,1}

\newcolumntype{V}{>{$\vcenter\bgroup\hbox\bgroup}c<{\egroup\egroup$}}

\graphicspath{{./imgs/}}

\cvprfinaltrue

\def\cvprPaperID{BICICCV-015}

\ifcvprfinal\fi
\begin{document}

\title{Synthesising Wider Field Images from Narrow-Field Retinal Video Acquired Using a Low-Cost Direct Ophthalmoscope (Arclight) Attached to a Smartphone}

\author{
\begin{tabular}{cccc}
  Keylor Daniel Chaves Viquez$^a$ &  Ognjen Arandjelovi\'c$^{\dag a}$  &  Andrew Blaikie$^b$ & In Ae Hwang$^b$\\
\end{tabular}\\
\begin{tabular}{cccc}
  $^a$School of Computer Science & $^b$School of Medicine\\
\end{tabular}\\
University of St Andrews\\
Scotland, United Kingdom\\
$^\dag$\small\texttt{ognjen.arandjelovic@gmail.com}
}
\maketitle

\begin{abstract}
Access to low cost retinal imaging devices in low and middle income countries is limited, compromising progress in preventing needless blindness. The Arclight is a recently developed low-cost solar powered direct ophthalmoscope which can be attached to the camera of a smartphone to acquire retinal images and video. However, the acquired data is inherently limited by the optics of direct ophthalmoscopy, resulting in a narrow field of view with associated corneal reflections, limiting its usefulness. In this work we describe the first fully automatic method utilizing videos acquired using the Arclight attached to a mobile phone camera to create wider view, higher quality still images comparable with images obtained using much more expensive and bulky dedicated traditional retinal cameras.
\end{abstract}

\section{Introduction}\label{s:intro}
It is estimated that some 285 million people in the world visually impaired, with the majority of cases potentially preventable or treatable if diagnosed promptly \cite{BlaiSandTuteWill+2016}. Unfortunately the burden of disease is greatest in low and middle income countries (East Asia, Latin America, and Africa) where access to diagnostic devices presents a major challenge \cite{GilbWormFielDeor+2015,BastHenn2012}. Observation of the retina by an eye care professional using an ophthalmoscope or through acquiring a video or photograph allows screening for key blinding diseases such as diabetic retinopathy, glaucoma and retinopathy of prematurity \cite{BlaiSandTuteWill+2016}.

\begin{figure}
  \centering
  \includegraphics[width=0.99\columnwidth]{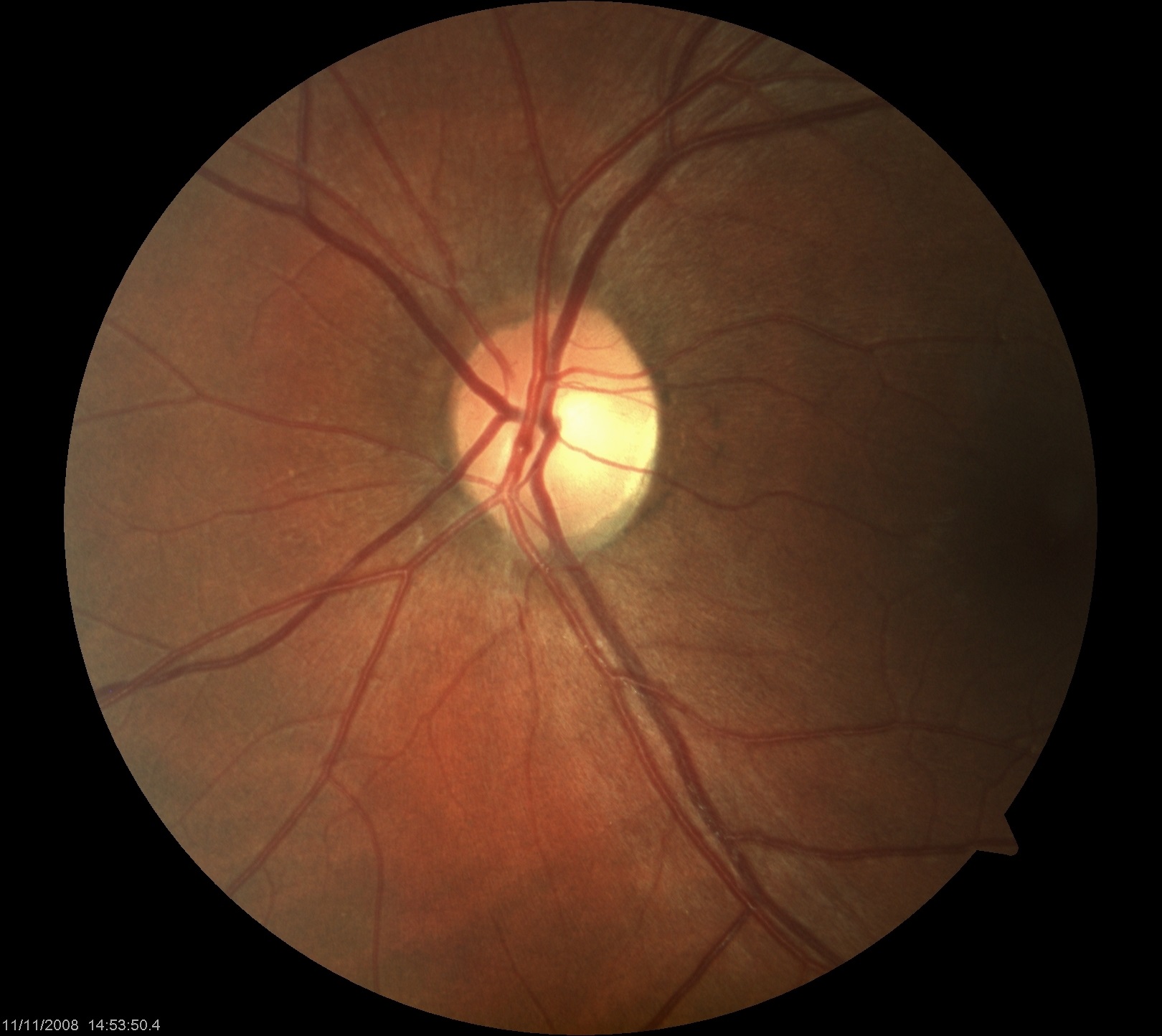}
  \caption{ High quality images of the eye fundus acquired using a traditional, expensive, and bulky tabletop unit. }
  \label{f:hq}
\end{figure}

However, one of the key barriers to fundal screening in low and middle income countries (LMICs) emerges from the lack of access to affordable retinal imaging equipment \cite{GilbWormFielDeor+2015}. The Arclight direct ophthalmoscope has been recently developed as a means of addressing this challenge \cite{BlaiSandTuteWill+2016}. The device, despite its simplified design and low cost, has been shown to be as effective as more expensive traditional devices \cite{LoweClelMgayFura+2015}. In addition, the Arclight can be attached to the camera of a mobile phone and acquire a digital recording of the fundus for the purposes of telemedicine. However, the captured images are intrinsically limited to a narrow field of view by the optics of direct ophthalmoscopy. Other limiting factors relating to the optics, sensor, and compression software of mobile phone cameras can further reduce the quality of the acquired data in comparison to more expensive dedicated traditional retinal cameras \cite{BlaiSandTuteWill+2016,GiarLiviJordBols+2014,BolsGiarBast2016}.

\section{Context and practical challenges}
Traditionally, retinal images (see Figure~\ref{f:hq}) are acquired using expensive and bulky tabletop units operated by trained technicians in a hospital setting  \cite{RussMaphTuraCost+2016}. Consequently such devices are impractical for use in low income areas where resources to buy and maintain traditional cameras are limited and robustness and portability is key for allowing the sharing and movement between inhospitable sites \cite{BlaiSandTuteWill+2016,KandSmitWrigHart2013}.

However, the availability of low cost yet powerful smartphones in LMICs has led to the development of `m-health' \cite{QianYamaHausAltm+2011} whereby mobile phones can play several roles: contacting and reminding patients to take medicines and attend appointments, transferring money to pay for health care, storing guidelines and diagnostic algorithms, acquiring and archiving audio and visual information from patients, analysing data, utilizisng GPS tracking, and transferring information wirelessly.  These approaches \cite{KandSmitWrigHart2013} have quickly been adopted in low resource health care settings where the portability and connectivity of smartphones \cite{Hull2008} has made them particularly valuable in telemedicine \cite{KandSmitWrigHart2013}, making it possible to reach patients and communities that currently receive suboptimal care due to geographical and financial barriers \cite{RussMaphTuraCost+2016}.

\subsection{The Arclight device}
\begin{figure}
  \centering
  \includegraphics[width=1\columnwidth]{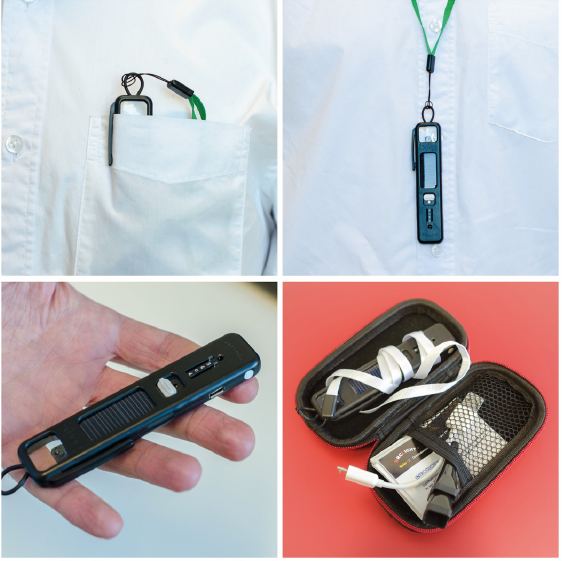}
  \caption{Arclight: a miniature, low-cost ophthalmoscope-otoscope. }
  \label{f:arcl}
\end{figure}

The Arclight is a highly portable (dimensions: $110$mm $\times$ $26$mm $\times$ $9$mm, mass: $18$g) low-cost (\pounds 5 if purchased in high volume\footnote{See \url{https://iapb.standardlist.org/get-arclight/}}) ophthalmoscope that has been developed in response to the need for affordable diagnostic eye care tools in LMICs \cite{TuteYoun2017}. Uniquely the device is illuminated by an LED allowing it to be powered by a slim lithium battery that is charged by an integrated solar panel reducing the need for expensive and hard to find consumables. Employing a small LED allows simplification of the design of the device reducing manufacturing costs and its overall size, see Figure~\ref{f:arcl} \cite{BlaiSandTuteWill+2016}. The novel features of the Arclight make it ideal for use in mobile clinics and remote low resource settings with limited access to power. Consequently over 10000 devices have been distributed, enabling thousands of health care workers worldwide to perform fundal examinations for the first time \cite{BlaiSandTuteWill+2016,TuteYoun2017,LoweClelMgayFura+2015}.

Studies have shown that the Arclight is just as accurate for screening for signs of diabetic retinopathy and glaucoma as more expensive traditional devices when used manually by an eye care professional \cite{BlaiSandTuteWill+2016}. However, one of the key inherent limitations of any direct ophthalmoscopy device is the narrow retinal field of view. Single still images are typically of limited utility but a sweeping video of a greater area of the fundus may offer greater clinical information, benefiting a telemedicine approach. Still, transferring video files is challenging in remote areas of a low income country where fast and reliable broadband internet may not be available. Synthesising a single wider field still image from multiple frames of a video fused together is a promising solution for overcoming many of the limitations of the optics of direct ophthalmoscopy as well as the data-heavy nature of video.

\begin{figure*}
  \centering
  \includegraphics[width=0.99\textwidth]{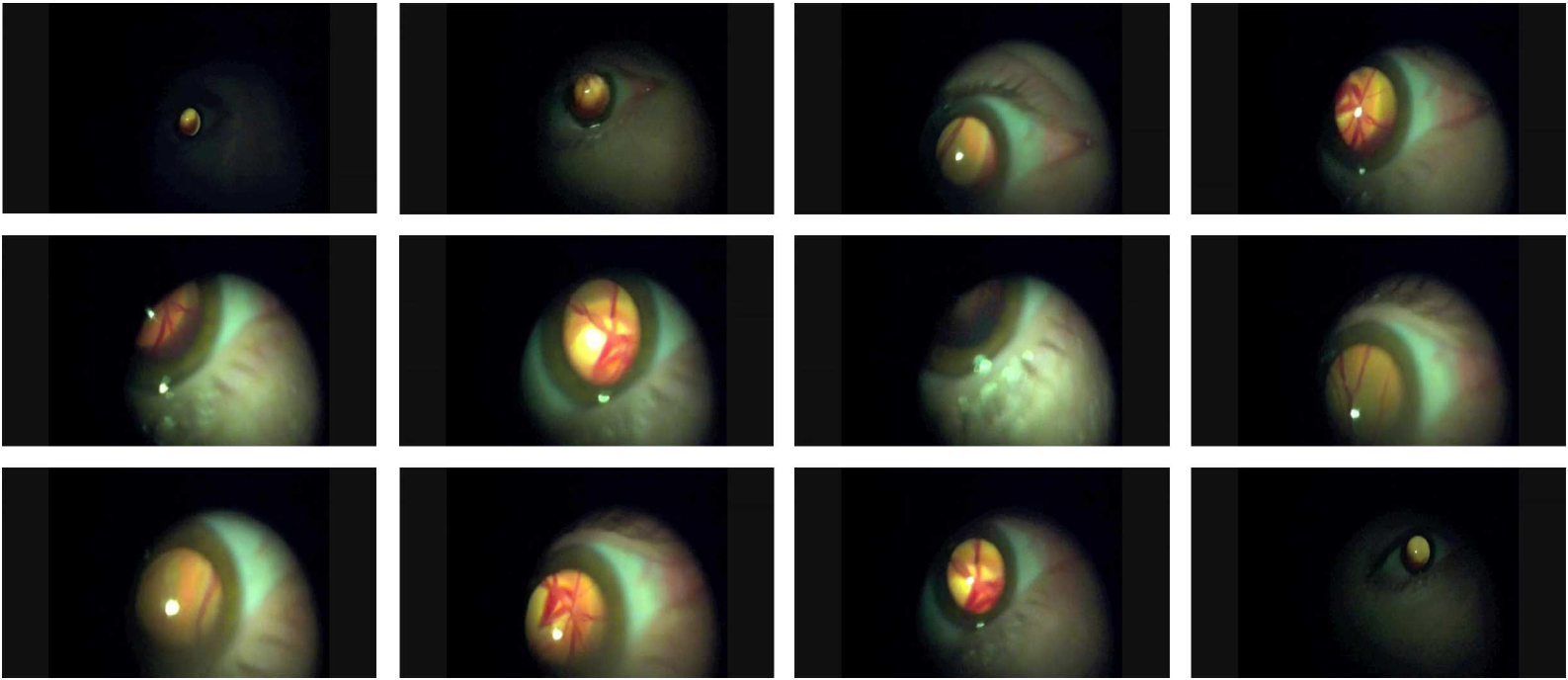}
  \caption{ Typical frames extracted from a video acquired using the Arclight device. Variable scale and brightness (which is also non-uniform), frequent occlusions of the region of interest, narrow field of view, and blur are just some of the notable challenges posed both to physicians and automated methods. }
  \label{f:inExample}
\end{figure*}

\section{Proposed method}
In this section we describe a novel method for processing narrow field retinal video acquired using the Arclight ophthalmoscope while attached to an Apple iPhone 6s. By fusing the information contained in different frames of a video we synthesise a wider field image of higher quality, which is easier to interpret clinically and more compact to transmit and share. We start with an overview of the method and then follow up with detailed descriptions of each of the processing steps.

\subsection{Overview}
As we noted earlier, the quality and useful information content of different frames in a video varies greatly. Hence, in order to ensure robustness, our algorithm starts by automatically identifying the highest quality frame. Tracking, and thereafter incremental image stitching, commences both in forward and backward directions in time. Each new frame is processed in order to remove confounding appearance content, processed to extract salient information, registered with the growing synthetic image (initially the single frame which which tracking begun), and stitched with it.

\subsection{Detection of the region of interest}
A crucial step in the pipeline outlined in the previous section concerns the detection of the region of interest (ROI) i.e.\ the part of the image which corresponds to the fundus and contains salient appearance content: the blood vessels and the optic disc. The parts of the frame which are outside of it are of no use for us and correspond to the sclera (the white, outer part of the eyeball), eyelids, etc.

As the frames in Figure~\ref{f:inExample} illustrate, the region of the interest is elliptical in shape, and bright. However, colour in the image can not be used reliably as a discriminative feature for segmentation as significant variation can be observed across different individuals (e.g.\ there are differences between individuals of Caucasian and African descent). Variation in location, ellipse shape and orientation, as well as scale should also be noted as significant challenges.

\begin{figure*}
  \centering
  \includegraphics[width=0.99\textwidth]{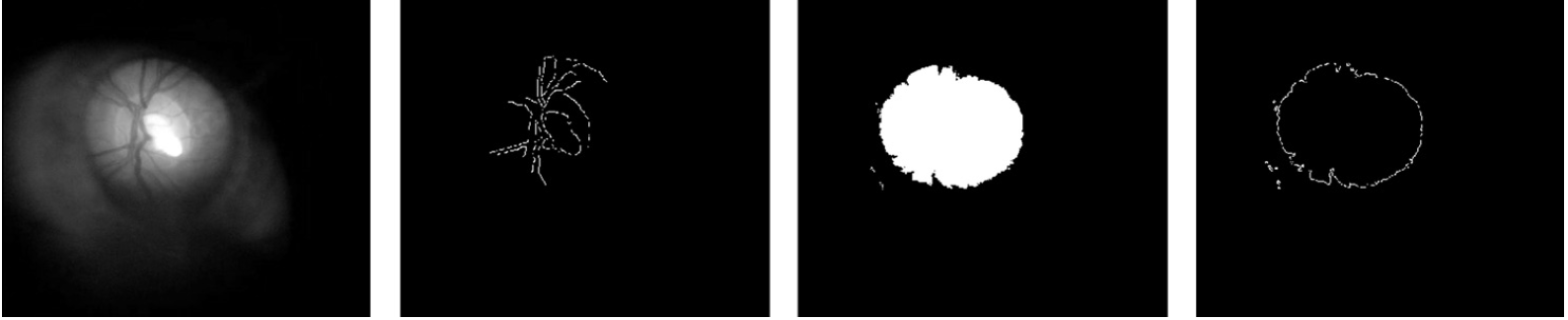}
  \caption{ Otsu's binary segmentation method based on automatic greyscale thresholding is used for the initial, rough localization of the region of interest. }
  \label{f:otsu}
\end{figure*}

Though the region of interest is elliptical in shape, observe that there is no sharp delineation, prohibiting the use of edge detection based approaches. Instead, we propose a two step method. In the first step, simple binary segmentation is performed using the well-known method first introduced by Otsu \cite{Otsu1979}. Here the frame is treated as a greyscale image, see Figure~\ref{f:otsu}. Then the contour of the largest region used to find the best elliptical fit. We achieve this using a genetic algorithm based approach. Our methodology is inspired by the work of Conn and Arandjelovi\'c who used ellipse fitting as the first stage in the task of localizing and segmenting ancient coins in real-world images~\cite{ConnAran2017}.

An ellipse embedded within a 2-dimensional Euclidean space is parameterized by five parameters. This can be readily seen from the implicit equation:
\begin{align}
  x^2 + k_{xy} x y + k_{yy} y^2 + k_x x + k_y y + k = 0.
  \label{e:ellipse}
\end{align}
The choice of the parameterization is crucial in making the most of our  genetic search strategy. In particular, the parameterization should be such that when encoded as a `chromosome' (in the context of a genetic algorithm), operations such as crossover, mutations, and others, are likely to effect an improvement in the fitness of a hypothesis. With this goal in mind, we parameterize an ellipse (hypothesis) using five points on its circumference and, following Conn and Arandjelovi\'c~\cite{ConnAran2017}, use a short, non-binary chromosome comprising the coordinates of these points. As argued by Conn and Arandjelovi\'c, by enforcing the indivisibility of coordinate values we ensure that the constraint imposed by two circumference points is retained during evolutionary operations, thereby achieving a higher chance of greater generational fitness improvement.

The fitness of a specific hypothesis is evaluated in a straightforward fashion. From chromosomes formed by concatenating the coordinates of five points on an ellipse's circumference -- $(x_a,y_a)$, $(x_b,y_b)$, $(x_c,y_c)$, $(x_d,y_d)$, and $(x_e,y_e)$ -- the parameters in \eqref{e:ellipse} can be obtained by solving a simple linear equation:
\begin{align}
  \left[
  \begin{array}{cccccc}
     x_a^2 & x_a y_a & y_a^2 &  x_a  & y_a & 1\\
     x_b^2 & x_b y_b & y_b^2 &  x_b  & y_b & 1\\
     x_c^2 & x_c y_c & y_c^2 &  x_c  & y_c & 1\\
     x_d^2 & x_d y_d & y_d^2 &  x_d  & y_d & 1\\
     x_e^2 & x_e y_e & y_e^2 &  x_e  & y_e & 1\\
  \end{array}\right]
  \left[\begin{array}{l}
    1\\ k_{xy}\\ k_{yy}\\ k_x\\ k_y\\ k\\
  \end{array}\right]
  =\left[\begin{array}{l}
    0\\0\\0\\0\\0\\
  \end{array}\right].
\end{align}

From this parameterization, the fitness of a hypothesis can be readily quantified by (i) uniformly sampling the circumference of the corresponding ellipse, and (ii) computing the number of samples which are located on edge pixels of the edge image. Equispaced samples along the circumference of the corresponding ellipse are readily generated using the canonical form, i.e.\ using the major and minor radii ($a$ and $b$) of the ellipse, its centre $(x_0, y_0)$, and the rotation angle $\theta$ relative to the coordinate system:
\begin{align}
  &a = \frac{-\sqrt{2 \psi \psi_1}}{k_{xy}^2 - 4k_{yy}}  &\text{and}&& b = \frac{-\sqrt{2 \psi \psi_2}}{k_{xy}^2 - 4k_{yy}},\\
  &x_0=\frac{2k_{yy} k_x  - k_{xy} k_y}{k_{xy}^2 - 4k_{yy}} &\text{and}&& y_0=\frac{2 k_y  - k_{xy} k_x}{k_{xy}^2 - 4k_{yy}},
\end{align}
where:
\begin{align}
 \psi=& k_y^2+ k_{yy}k_x^2 - k_{xy}k_x k_y + (k_{xy}^2 - 4k_{yy})k,\\
 \psi_1=& 1+k_{yy} + \sqrt{(1-k_{yy})^2+k_{xy}^2},\\
 \psi_2=& 1+k_{yy} - \sqrt{(1-k_{yy})^2+k_{xy}^2}.
\end{align}
Then, if there are $n_s$ samples, and the samples are $\left\{ (x_1,y_1),(x_2,y_2),\ldots,(x_{n_s},y_{n_s})\right\}$, the fitness becomes:
\begin{align}
  \phi = \frac{ \sum_{i=1}^{n_s} E(x_i,y_i) }{ n_s},
  \label{e:fitness}
\end{align}
where $E(x,y)$ is the value of the pixel $(x,y)$ in the edge image $\mathbf{E}$.

\begin{figure*}
  \centering
  \includegraphics[width=2\columnwidth,height=1\columnwidth]{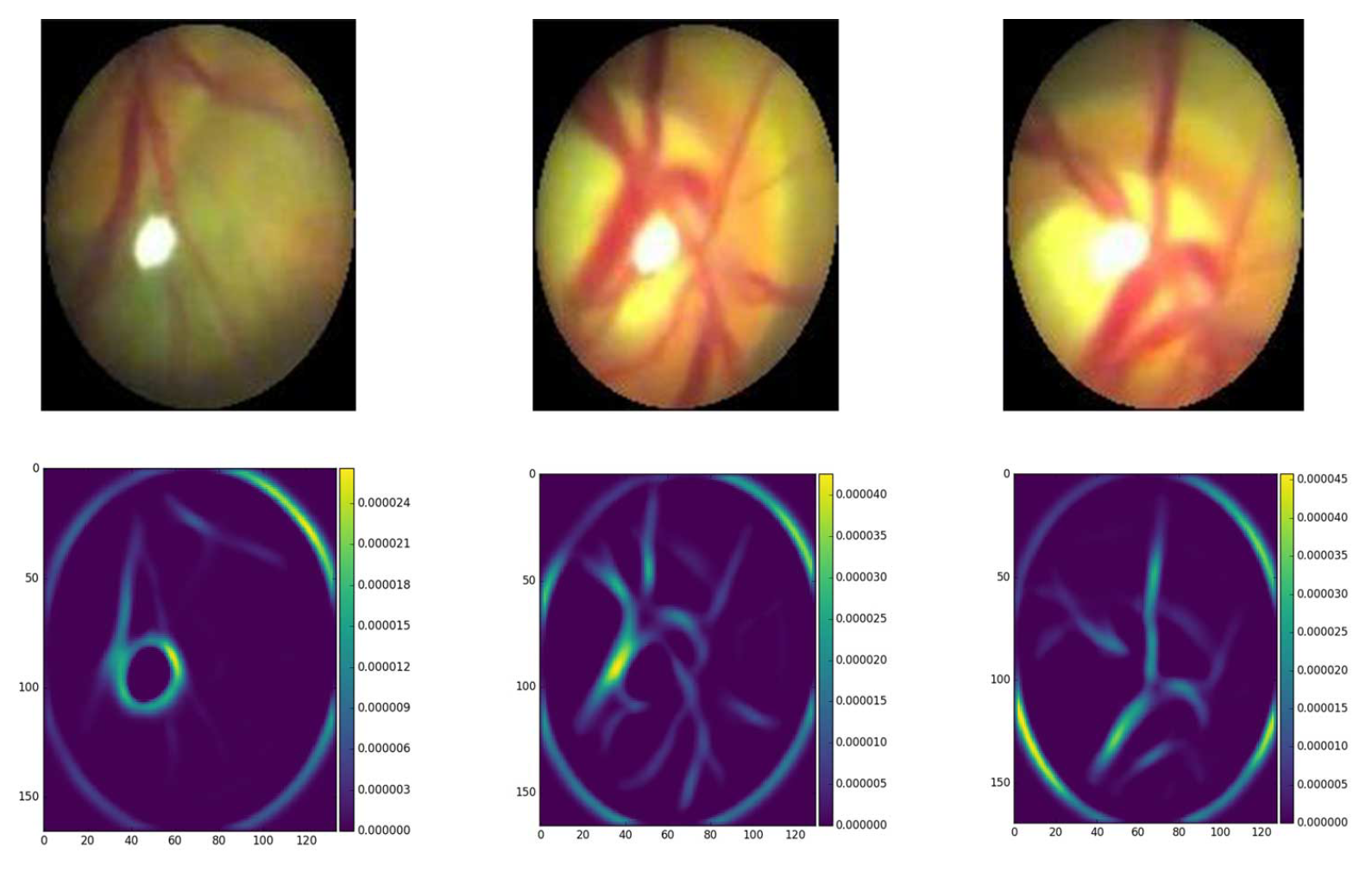}
  \caption{ Examples of extracted vessel structures. Note that here only the raw output is shown: artefacts at the borders of the region of interest and the glare and removed thereafter. }
  \label{f:frangi}
\end{figure*}

\subsection{Glare detection and removal}
As mentioned earlier, one of the challenges with the input videos is the presence of significant glare. It is of critical importance that the areas of the fundus occluded by glare are detected so as not to confound the tracking process, as well as so that the relevant regions can be filled-in by exploiting information from other frames in which they are not occluded.

Glare areas are for practical purposes of unpredictable shape, often but not universally present, and their location varies. All of these characteristics make their detection difficult. Another challenge which emerges in our specific application concerns the appearance of the optic disk which is usually brighter than the rest of the fundus, and could be mistaken for glare. Yet, the appearance of the optic disk is a crucial structure of interest to ophthalmologists, used to diagnose glaucoma and a number of other conditions \cite{KiagKherGichDamj+2013}.

\begin{figure}
  \centering
  \includegraphics[width=1\columnwidth,height=1.1\columnwidth]{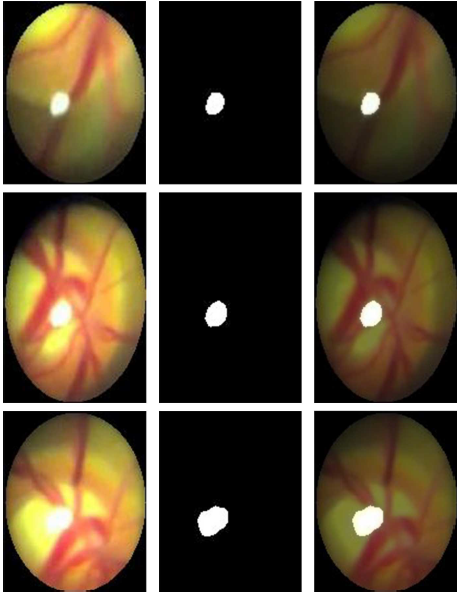}
  \caption{ Glare detection result examples: original images (left), and the corresponding binary glare masks (centre) and images with glare removed (right). }
  \label{f:glare}
\end{figure}

Our approach to detecting glare is inspired by the work of Lange \cite{Lang2005} on images of the uterine cervix. Broadly speaking, we detect glare areas as those regions of the image which exhibit high contrast or saturation, and are approximately white. Formally, the criterion is based on the following local measure $g(x,y)$:
\begin{align}
   g(x,y)=\min(R_{x,y},G_{x,y},B_{x,y}) - \frac{\min(R_{x,y},G_{x,y},B_{x,y})}{\max(R_{x,y},G_{x,y},B_{x,y})}
\end{align}
where $R_{x,y}$, $G_{x,y}$, and $B_{x,y}$ are respectively the red, green, and blue colour components of the pixel at the locus $(x,y)$. After computing $g(x,y)$ for all image loci the corresponding feature image is smoothed using the Alternating Sequential Filter (a sequence of morphological closing and opening operations), and the binary mask $\mathbf{M}$ computed with $M_{x,y} = 1$ iff $G_{x,y}$ exceeds a predefined threshold, and $M_{x,y} = 0$ otherwise. Examples are shown in Figure~\ref{f:glare}.

\subsection{Salient information extraction}
Having removed confounding image content (glare, background, etc.) the processed frames are ready for the main part of our algorithms with concerns the tracking of salient structures present in the fundus, and the subsequent stitching. However, notwithstanding the significant processing described thus far, these tasks are still far from straightforward. For example, note that neither simple appearance nor any of the commonly used local features can be used for tracking (or, equivalently, registration). Raw appearance changes greatly due to global illumination changes, as well as local changes caused by the inherent optics of the eye. Local features -- such as appearance based SIFT \cite{Lowe2004} and SURF \cite{BayEssTuytGool2008}, or edge based ones \cite{Aran2012f} -- can also not be reliably detected as the appearance of the fundus does not contain corner-like loci.

Our approach is specifically targeted at the type of appearance content of interest in fundal images, namely blood vessels. These are line like features which form complex and characteristic structures, and by virtue of this allow different frames to be mutually registered.

We extract characteristic vessel based features using the method introduced by Frangi \textit{et al}.\ \cite{FranNiesVincVier1998}. Their so-called vesselness filter, first proposed for the use on 3D MRI data but later successfully employed in a number of different applications \cite{GhiaAranBendMald2013a,GhiaAranBendMald2013,GhiaAranBendMald2014a}, extracts tubular structures from an image. For a 2D image consider the two eigenvalues $\lambda_1$ and $\lambda_2$ of the Hessian matrix computed at a certain image locus and at a particular scale. Without loss of generality let us also assume that $|\lambda_1| \leq |\lambda_2|$. The two key values used to quantify how tubular the local structure at this scale is are $\mathcal{R}_\mathcal{A} = |\lambda_1|/|\lambda_2|$ and $\mathcal{S} = \sqrt{\lambda_1^2 + \lambda_1^2}$. The former of these measures the degree of local 'blobiness'. If the local appearance is blob-like, the Hessian is approximately isotropic and $|\lambda_1|\approx|\lambda_2|$ making $\mathcal{R}_\mathcal{A}$ close to 1. For a tubular structure $\mathcal{R}_\mathcal{A}$ should be small. On the other hand, $\mathcal{S}$ ensures that there is sufficient local information content at all: in nearly uniform regions, both eigenvalues of the corresponding Hessian will have small values. For a particular scale of image analysis $s$, the two measures, $\mathcal{R}_\mathcal{A}$ and $\mathcal{S}$, are then unified into a single vesselness measure:
{\small\begin{align}
  \mathcal{V}(s) =
    \begin{cases}
      0 &~ \text{if } \lambda_2 > 0\\
      (1-e^{-\frac{\mathcal{R}_\mathcal{B}}{2\beta^2}}) \times (1-e^{-\frac{\mathcal{S}}{2c^2}}) &~ \text{otherwise},
    \end{cases}
\end{align}}
where $\beta$ and $c$ are the parameters that control the sensitivity of the filter to $\mathcal{R}_\mathcal{A}$ and $\mathcal{S}$. Finally, if an image is analyzed across scales from $s_{min}$ to $s_{max}$, the vesselness of a particular image locus can be computed as the maximal vesselness across the range:
{\small\begin{align}
  \mathcal{V}_0 = \max_{s_{min} \leq s \leq s_{max}} \mathcal{V}(s)
\end{align}}
We empirically set the parameter values to $\beta= 0.75$ and $c = 15$, and process frames at scales $s=3,4,5$. Examples of typical results are shown in Figure~\ref{f:frangi}.

\subsubsection{Algorithm initialization}
Recall that our algorithm starts the tracking (and thus the stitching process) from an automatically frame detected as the most reliable one i.e.\ one that contains sufficient salient information content to facilitate robust registration with the subsequent frames, and free of motion or out of focus blur. The vessel extraction method we just described is used to this end. In particular, following the computation of the vesselness image, the richness of salient information content is quantified by computing the entropy of this image. Input frames which do not contain many blood vessels, or which are corrupted by blur, will not produce many vessel detections, and thus result in low entropy vesselness images. Therefore, the frame with the highest entropy of the corresponding vesselness image is selected as the starting frame from which tracking commences, and as the initial synthetic image which is subsequently extended and improved though merging with newly processed frames.

\subsection{Image registration}
Recall that our algorithm incrementally expands the synthetic image of the fundus by expanding it by (stitching with) newly processed frames. The first step in the stitching process concerns the registration of a frame with the synthetic image. This is achieved by using the extracted vessel feature images. Specifically, we adopt the use of normalized cross-correlation as the hypothesis matching criterion \cite{AranPhamVenk2015c}, and exhaustively (though in a coarse to fine fashion, to speed up the process) explore the space of different scaling and translation parameters. The combination of the parameters which results in the highest normalized cross-correlation is accepted as the correct hypothesis but only if matching goodness exceeds that of the second best hypothesis (the second highest local normalized cross-correlation maximum) by a significant enough margin (herein we used a factor of 1.5). If this condition is not satisfied (which can happen, for example, when the newly acquired frame does not contain enough information to allow confident registration) the frame is discarded as unreliable.

\subsection{Incremental stitching}
Having registered a new frame with the growing synthetic image, the last step of our algorithm concerns the merging of the two images and with it the expansion of the synthetic result. There are several factors which make the process difficult. Firstly and as we noted before, there may be a significant illumination discrepancy between the two images. Moreover, a gradual darkening of the region of interest in the proximity of its boundary can be readily observed; note that this is a feature of the imaging process, rather than an artefact introduced by any of the steps of our algorithm. Lastly, the regions corresponding to glare areas, void of useful information, should be taken into account.

\begin{figure*}
  \centering
  \includegraphics[width=1.85\columnwidth,height=.57\columnwidth]{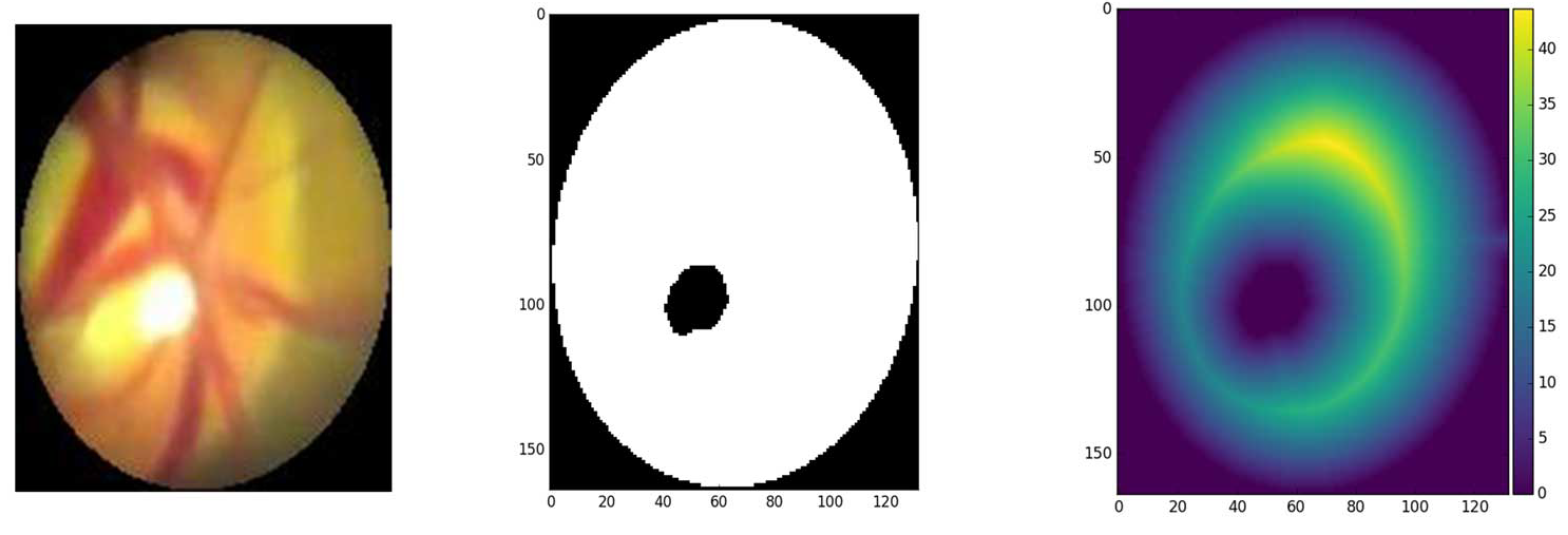}
  \caption{ Original frame (left), region of interest (after the removal of the background and the glare), and the corresponding distance transform used for adaptive pair-wise image merging. }
  \label{f:dt}
\end{figure*}

Ideally, the result of the merging process should not only produce a seamless output image, spatially expanded (in general), but also leverage information content from both input images to increase the quality of information in the overlapping areas too. To achieve this while at the same addressing the challenges summarized above, we adopt an adaptive weighting strategy whereby the relative contributions of the two input images are adjusted depending on the particular locus under the consideration. To motivate out approach intuitively, the idea is to use preferentially data from an image in which a particular location is more central (relative to the region of interest) i.e.\ where the optical setup of mini ophthalmoscopies and the eye itself, produce better quality data (higher signal to noise ratio). We formalize this through the use of the distance transform \cite{Borg1984,Aran2012b}. Specifically, after a linear illumination transform of brightness which normalizes global changes, for each pixel in an image we compute its approximate distance from the boundaries of the region of interest (the outer boundary and, if present, the boundary of the glare region). Then, when combining two corresponding pixels their relative contributions are computed by considering the inverse value of their distances from the boundaries in the original images, i.e.\ the output value of a RGB channel $c_{x,y}$ becomes:
\begin{align}
  c_{x,y}= \frac{w_f c^f_{x,y} + w_s c^s_{x,y}}{w_f w_s}
\end{align}
where $c^f_{x,y}$ is the value of the pixel in a new frame and $c^s_{x,y}$ in the synthetic image, and:
\begin{align}
  w_f = 1/d^f_{x,y} && w_s = 1/d^s_{x,y}
\end{align}
where $d^f_{x,y}$ and $d^s_{x,y}$ are the corresponding distance transform values (see Figure~\ref{f:dt}). The process can be therefore described as a form of adaptive linear fusion \cite{AranCipo2006a,AranHammCipo2010,Aran2016d}.

\begin{figure*}[!ht]
  \centering
  \includegraphics[width=0.99\textwidth]{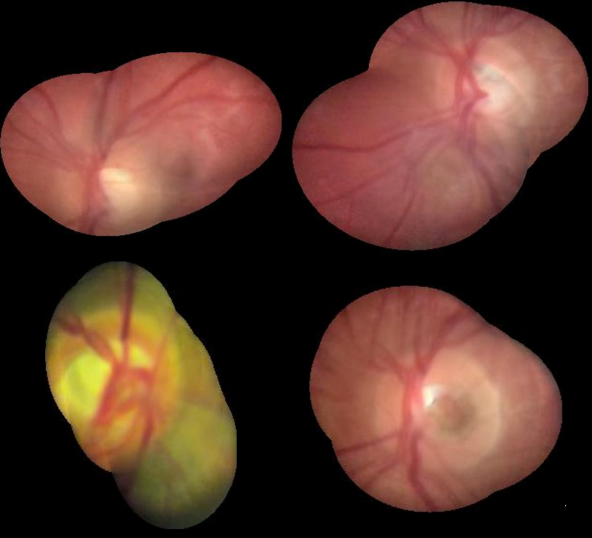}
  \caption{ Examples of synthetically generated images. }
  \label{f:res1}
\end{figure*}

\begin{figure*}[!ht]
  \centering
  \includegraphics[width=0.99\textwidth]{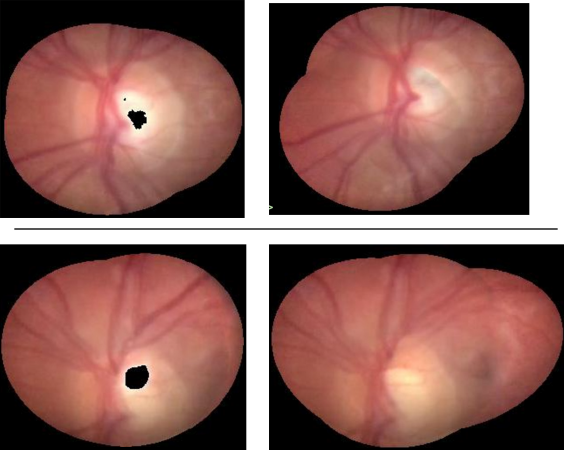}
  \caption{ Examples of image fusion with the specific emphasis on the handling of glare. }
  \label{f:res2}
\end{figure*}

\section{Evaluation}
The data corpus we used in this work was collected from healthy volunteers. The acquisition process was entirely unaffected and unguided by us, guaranteeing realistic video input as would be collected in the field. Thus, we found that the sequences were highly variable in length, lasting from approximately 20 up to 90 seconds. The majority of the sequences start with a distant view of the eye (and hence little useful fundal information), and progressively close in onto the areas of actual interest such as blood vessels and the optic disc. Major motion and out of focus blur can be observed in this early phase. Once a section of the retina is clear the acquisition becomes more controlled, exhibiting purposeful motion in an effort to capture as much of the fundus as possible. Only a small section can be seen at any given moment due to the narrow field of view of direct ophthalmoscopy devices. Throughout this process blur continues to pose problems, with the retina easily going out of focus, drastic changes in the direction of motion occurring, and the imaging stopping unexpectedly. Considering the nature of our data set, it was not possible to obtain objective ground truth high quality images using a traditional ophthalmoscope. Hence, at this stage we limited ourselves to a qualitative assessment of the results. Representative examples of synthetic images and the effects of specific challenges are shown in Figures~\ref{f:res1} and~\ref{f:res2}.

\section{Summary and conclusions}\label{s:summary}
The Arclight is a recently developed low-cost direct ophthalmoscope that offers great potential for the prevention of blindness amongst those living in low and middle income countries. However, currently this potential is limited by the inherent optical limitations of direct ophthalmoscopy, most notably the narrow field of view. In this paper we described the first method that uses this low quality data to create a synthetic, higher quality, wider view image comparable to one acquired using expensive and bulky traditional retinal cameras.

\balance


{\small
\bibliographystyle{ieee}
\bibliography{./my_bibliography}

\begin{thebibliography}{10}\itemsep=-1pt

\bibitem{Aran2012b}
O.~Arandjelovi{\'c}.
\newblock Colour invariants under a non-linear photometric camera model and
  their application to face recognition from video.
\newblock {\em Pattern Recognition}, 45(7):2499--2509, 2012.

\bibitem{Aran2012f}
O.~Arandjelovi{\'c}.
\newblock Object matching using boundary descriptors.
\newblock {\em In Proc.\ British Machine Vision Conference}, 2012.
\newblock {DOI:~10.5244/C.26.85}.

\bibitem{Aran2016d}
O.~Arandjelovi{\'c}.
\newblock Weighted linear fusion of multimodal data -- a reasonable baseline?
\newblock {\em In Proc.\ ACM Conference on Multimedia}, pages 851--857, 2016.

\bibitem{AranCipo2006a}
O.~Arandjelovi{\'c} and R.~Cipolla.
\newblock A new look at filtering techniques for illumination invariance in
  automatic face recognition.
\newblock {\em In Proc.\ IEEE International Conference on Automatic Face and
  Gesture Recognition}, pages 449--454, 2006.

\bibitem{AranHammCipo2010}
O.~Arandjelovi{\'c}, R.~I. Hammoud, and R.~Cipolla.
\newblock Thermal and reflectance based personal identification methodology in
  challenging variable illuminations.
\newblock {\em Pattern Recognition}, 43(5):1801--1813, 2010.

\bibitem{AranPhamVenk2015c}
O.~Arandjelovi{\'c}, D.~Pham, and S.~Venkatesh.
\newblock Efficient and accurate set-based registration of time-separated
  aerial images.
\newblock {\em Pattern Recognition}, 48(11):3466--3476, 2015.

\bibitem{BastHenn2012}
A.~Bastawrous and B.~D. Hennig.
\newblock The global inverse care law: a distorted map of blindness.
\newblock {\em British Journal of Ophthalmology}, 96(10):1357--1358, 2012.

\bibitem{BayEssTuytGool2008}
H.~Bay, A.~Ess, T.~Tuytelaars, and L.~V. Gool.
\newblock {SURF}: Speeded up robust features.
\newblock {\em Computer Vision and Image Understanding}, 110(3):346--359, 2008.

\bibitem{BlaiSandTuteWill+2016}
A.~Blaikie, J.~Sandford-Smith, S.~Y. Tuteja, C.~D. Williams, and
  C.~O'Callaghan.
\newblock Arclight: a pocket ophthalmoscope for the 21st century.
\newblock {\em The British Medical Journal}, 355:i6637, 2016.

\bibitem{BolsGiarBast2016}
N.~M. Bolster, M.~E. Giardini, and A.~Bastawrous.
\newblock The diabetic retinopathy screening workflow.
\newblock {\em Journal of Diabetes Science and Technology}, 10(2):318--324,
  2016.

\bibitem{Borg1984}
G.~Borgefors.
\newblock Distance transforms in arbitrary dimensions.
\newblock {\em Computer Vision, Graphics and Image Processing}, pages 321--345,
  1984.

\bibitem{ConnAran2017}
B.~Conn and O.~Arandjelovi{\'c}.
\newblock Towards computer vision based ancient coin recognition in the wild --
  automatic reliable image preprocessing and normalization.
\newblock {\em In Proc.\ IEEE International Joint Conference on Neural
  Networks}, pages 1457--1464, 2017.

\bibitem{FranNiesVincVier1998}
A.~F. Frangi, W.~J. Niessen, K.~L. Vincken, and M.~A. Viergever.
\newblock Multiscale vessel enhancement filtering.
\newblock {\em Medical Image Computing and Computer-Assisted Intervention},
  pages 130--137, 1998.

\bibitem{GhiaAranBendMald2013a}
R.~S. Ghiass, O.~Arandjelovi{\'c}, A.~Bendada, and X.~Maldague.
\newblock Illumination-invariant face recognition from a single image across
  extreme pose using a dual dimension {AAM} ensemble in the thermal infrared
  spectrum.
\newblock {\em In Proc.\ IEEE International Joint Conference on Neural
  Networks}, pages 2781--2790, 2013.

\bibitem{GhiaAranBendMald2013}
R.~S. Ghiass, O.~Arandjelovi{\'c}, A.~Bendada, and X.~Maldague.
\newblock Vesselness features and the inverse compositional {AAM} for robust
  face recognition using thermal {IR}.
\newblock {\em In Proc.\ AAAI Conference on Artificial Intelligence},
  1(0):357--364, 2013.

\bibitem{GhiaAranBendMald2014a}
R.~S. Ghiass, O.~Arandjelovi{\'c}, A.~Bendada, and X.~Maldague.
\newblock A unified framework for thermal {IR}-based face recognition.
\newblock {\em In Proc.\ International Conference on Neural Information
  Processing}, II:335--343, 2014.

\bibitem{GiarLiviJordBols+2014}
M.~E. Giardini, I.~A.~T. Livingstone, S.~Jordan, N.~M. Bolster, T.~Peto,
  M.~Burton, and A.~Bastawrous.
\newblock A smartphone based ophthalmoscope.
\newblock {\em In Proc.\ International Conference of the IEEE Engineering in
  Medicine and Biology Society}, pages 2177--2180, 2014.

\bibitem{GilbWormFielDeor+2015}
C.~Gilbert, R.~Wormald, A.~Fielder, A.~Deorari, L.~C. Zepeda-Romero, G.~Quinn,
  A.~Vinekar, A.~Zin, and B.~Darlow.
\newblock Potential for a paradigm change in the detection of retinopathy of
  prematurity requiring treatment.
\newblock {\em Archives of Disease in Childhood -- Fetal and Neonatal Edition},
  101(1):6--9, 2015.

\bibitem{Hull2008}
C.~Hull.
\newblock Digital imaging and screening for diabetic retinopathy.
\newblock {\em Optometry Today}, pages 28--35, 2008.

\bibitem{KandSmitWrigHart2013}
Y.~Kandasamy, R.~Smith, I.~Wright, and L.~Hartley.
\newblock Use of digital retinal imaging in screening for retinopathy of
  prematurity.
\newblock {\em Journal of Paediatrics and Child Health}, 49(1), 2013.

\bibitem{KiagKherGichDamj+2013}
D.~Kiage, I.~N. Kherani, S.~Gichuhi, K.~F. Damji, and M.~Nyenze.
\newblock The {M}uranga teleophthalmology study: comparison of virtual
  (teleglaucoma) with in-person clinical assessment to diagnose glaucoma.
\newblock {\em Middle East African Journal of Ophthalmology}, 20(2):150--157,
  2013.

\bibitem{Lang2005}
H.~Lange.
\newblock Automatic glare removal in reflectance imagery of the uterine cervix.
\newblock {\em In Proc.\ SPIE}, 5747:2183--2192, 2005.

\bibitem{Lowe2004}
D.~G. Lowe.
\newblock Distinctive image features from scale-invariant keypoints.
\newblock {\em International Journal of Computer Vision}, 60(2):91--110, 2003.

\bibitem{LoweClelMgayFura+2015}
J.~Lowe, C.~R. Cleland, E.~Mgaya, G.~Furahini, C.~E. Gilbert, M.~J. Burton, and
  H.~Philippin.
\newblock The {A}rclight ophthalmoscope: a reliable low-cost alternative to the
  standard direct ophthalmoscope.
\newblock {\em Journal of Ophthalmology}, 2015, 2015.

\bibitem{Otsu1979}
N.~Otsu.
\newblock A threshold selection method from gray-level histograms.
\newblock {\em IEEE Transactions on Systems, Man, and Cybernetics},
  9(1):62--66, 1979.

\bibitem{QianYamaHausAltm+2011}
C.~Z. Qiang, M.~Yamamichi, V.~Hausman, D.~Altman, and {IS Unit}.
\newblock Mobile applications for the health sector.
\newblock {\em Washington: World Bank}, 2011.

\bibitem{RussMaphTuraCost+2016}
A.~Russo, W.~Mapham, R.~Turano, C.~Costagliola, F.~Morescalchi, N.~Scaroni, and
  F.~Semeraro.
\newblock Comparison of smartphone ophthalmoscopy with slit-lamp biomicroscopy
  for grading vertical cup-to-disc ratio.
\newblock {\em Journal of Glaucoma}, 25(9):e777--781, 2016.

\bibitem{TuteYoun2017}
S.~Tuteja and T.~Young-Zvandasara.
\newblock The {A}rclight: A pocket ophthalmoscope to revitalise undergraduate
  teaching?
\newblock {\em Eye News UK}, 23(4), 2017.

\end{thebibliography}
}

\end{document}